\documentclass[nohyperref]{article}

\usepackage{microtype}
\usepackage{graphicx}
\usepackage{subfigure}
\usepackage{booktabs} 
\usepackage{enumitem}
\usepackage{xcolor}
\usepackage[backref=page]{hyperref}

\usepackage[accepted]{pre-training}

\usepackage{amsmath}
\usepackage{amssymb}
\usepackage{mathtools}
\usepackage{amsthm}
\usepackage{tabularx}

\usepackage[capitalize,noabbrev]{cleveref}

\theoremstyle{plain}

\theoremstyle{definition}

\theoremstyle{remark}

\usepackage[textsize=tiny]{todonotes}

\newcommand{\std}[1]{{\scriptsize{$\pm$#1}}}

\newcommand{\f}{a}

\newcommand{\domain}{\mathcal{D}}

\newcommand{\w}{\mathbf{w}}

\icmltitlerunning{Pretraining a Neural Network before Knowing Its Architecture}

\begin{document}

\twocolumn[
\icmltitle{Pretraining a Neural Network before Knowing Its Architecture}



\icmlsetsymbol{equal}{*}

\begin{icmlauthorlist}
\icmlauthor{Boris Knyazev}{yyy}
\end{icmlauthorlist}

\icmlaffiliation{yyy}{Samsung - SAIT AI Lab, Montreal, Canada}

\icmlcorrespondingauthor{Boris Knyazev's homepage:}{\url{http://bknyaz.github.io/}}

\icmlkeywords{Machine Learning, ICML}

\vskip 0.3in
]



\printAffiliationsAndNotice{} 

\begin{abstract}
Training large neural networks is possible by training a smaller \textit{hypernetwork} that predicts parameters for the large ones. A recently released Graph HyperNetwork (GHN) trained this way on one million smaller ImageNet architectures is able to predict parameters for large unseen networks such as ResNet-50. While networks with predicted parameters lose performance on the source task, the predicted parameters have been found useful for fine-tuning on other tasks. We study if fine-tuning based on the same GHN is still useful on novel strong architectures that were published after the GHN had been trained. We found that for recent architectures such as ConvNeXt, GHN initialization becomes less useful than for ResNet-50. One potential reason is the increased distribution shift of novel architectures from those used to train the GHN. We also found that the predicted parameters lack the diversity necessary to successfully fine-tune parameters with gradient descent. We alleviate this limitation by applying simple post-processing techniques to predicted parameters before fine-tuning them on a target task and improve fine-tuning of ResNet-50 and ConvNeXt.\looseness-1
\end{abstract}

\begin{figure}[t!]
\begin{center}
    \includegraphics[width=0.95\columnwidth]{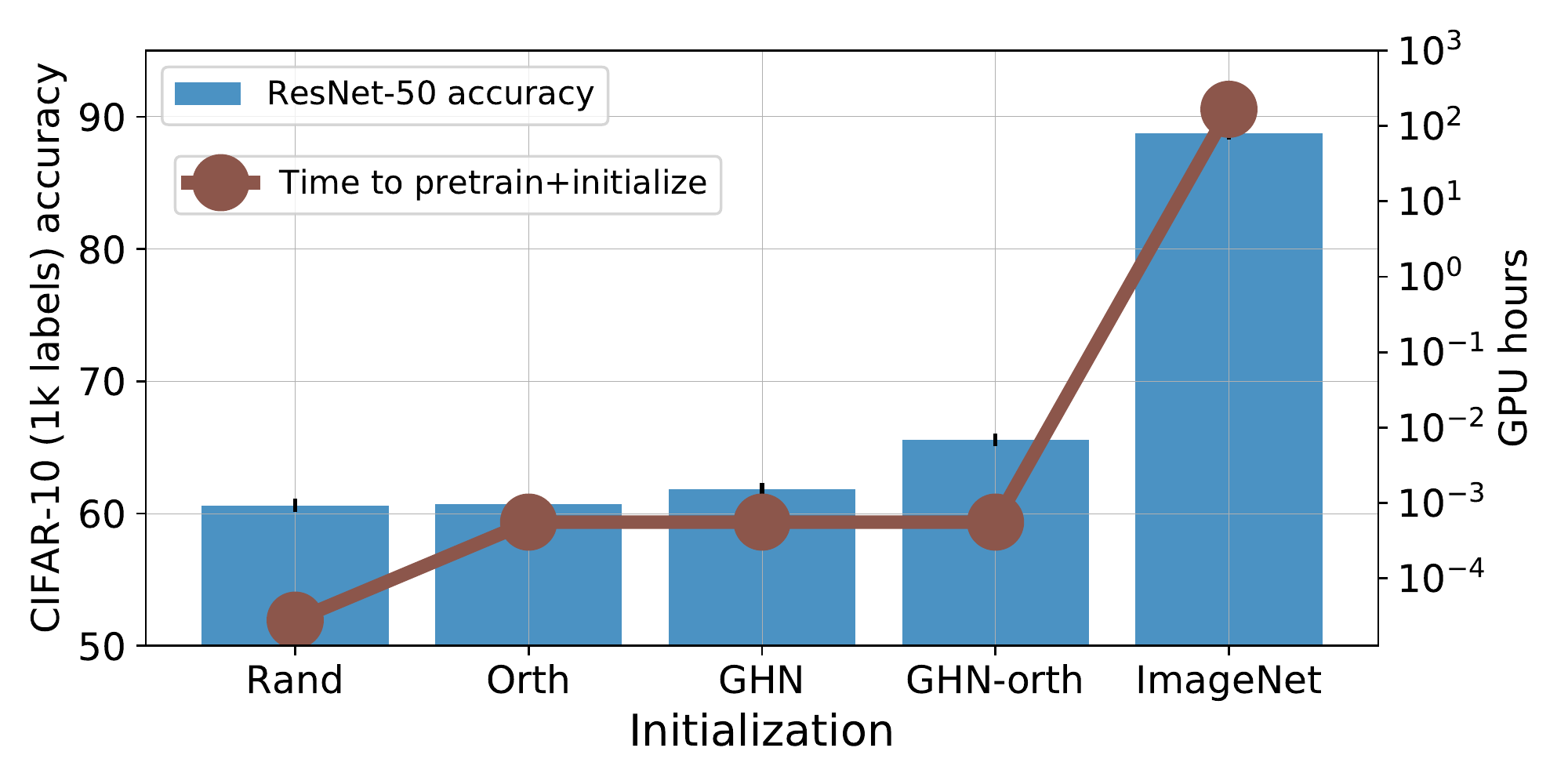}
    \vspace{-5pt}
    \caption{Motivating example showing a large amount of GPU hours (note the log scale) required to pretrain a novel architecture on ImageNet-1k to use it for initialization. To reduce the computational burden arising in practice when having own large pretraining dataset, we study the strategy of reusing an already trained Graph HyperNetwork (GHN)~\cite{knyazev2021parameter} for novel architectures, such as ConvNeXt~\cite{liu2022convnet}, even if the architectures are discovered after the GHN was trained. {GHN-orth} is the proposed initialization. GPU hours are based on~\citet{knyazev2021parameter}. See Table~\ref{tab:results} for the reported accuracies.}
    \label{fig:results}
\end{center}
\vskip -0.2in    
\end{figure}

\section{Graph HyperNetworks for Initialization}

Initialization of deep neural nets is critical to make them converge fast and to a generalizable solution~\cite{glorot2010understanding,he2015delving}. 
When training a neural net on small training data, an effective approach to initialize it is to pretrain it on a large dataset~\cite{huh2016makes,kolesnikov2020big}, such as ImageNet~\cite{russakovsky2015imagenet}.
The architectures of neural nets keep evolving due to efforts of humans~\cite{dosovitskiy2020image,liu2022convnet} and neural architecture search~\cite{elsken2019neural}. So practitioners often need to rerun a costly pretraining procedure on their large in-house data\footnote{E.g. Google's JFT-300M and Facebook's IG-1B-Targeted.} for every new  architecture discovered by the community to initialize it this way (Figure~\ref{fig:results}).\looseness-1

In the long term, a more efficient approach to initialize neural nets may be to train a Graph HyperNetwork (GHN)~\cite{zhang2018graph, knyazev2021parameter} that can predict parameters for different architectures, including those yet to be discovered.
GHN $H_{\domain}$ parameterized by $\theta$ needs to be trained only once on a large pretraining dataset ${\domain}$ (e.g. ImageNet). It can then predict parameters $\w$ in fractions of a second for arbitrary\footnote{Any feedforward neural network architecture that can be represented as a directed acyclic graph (DAG) composed of the same primitive operations used during training GHNs.} architectures $\f$:
$\w=H_{\domain}(\f; \theta)$.
GHNs can predict parameters for much larger architectures than seen during training such as ResNet-50~\cite{he2016deep}.
While networks with predicted parameters lose performance on the source task $\domain$, the predicted parameters have been found useful as initialization for fine-tuning on other tasks~\cite{knyazev2021parameter}. Such an initialization compared favorably to random-based initialization methods~\cite{he2015delving}.

We study if initialization based on the already available trained GHN is still useful on novel strong architectures that were found after the GHN had been trained.
We consider a real use case by evaluating the released GHN of~\citet{knyazev2021parameter} on a recent ConvNeXt architecture~\cite{liu2022convnet}. We found that for ConvNeXt the parameters predicted by the GHN become less useful for initialization and fine-tuning than for earlier architectures such as ResNet-50.
One potential reason for that is an increased distribution shift of novel architectures from those used to train the GHN.
We also analyzed predicted parameters (Section~\ref{sec:analyze}) and found that when initializing networks with predicted parameters, fine-tuning performance can be improved by reducing the similarities of predicted parameters (Section~\ref{sec:postproc}).
The code to reproduce our results is added to the official GHN repository \url{https://github.com/facebookresearch/ppuda}.

\begin{figure}[t]
\vskip 0.1in
\begin{center}
\setlength{\tabcolsep}{0pt}
\begin{small}
\begin{tabular}{ccc}
        first layer & shallow layer & deeper layer \\
         {\includegraphics[width=0.33\columnwidth]{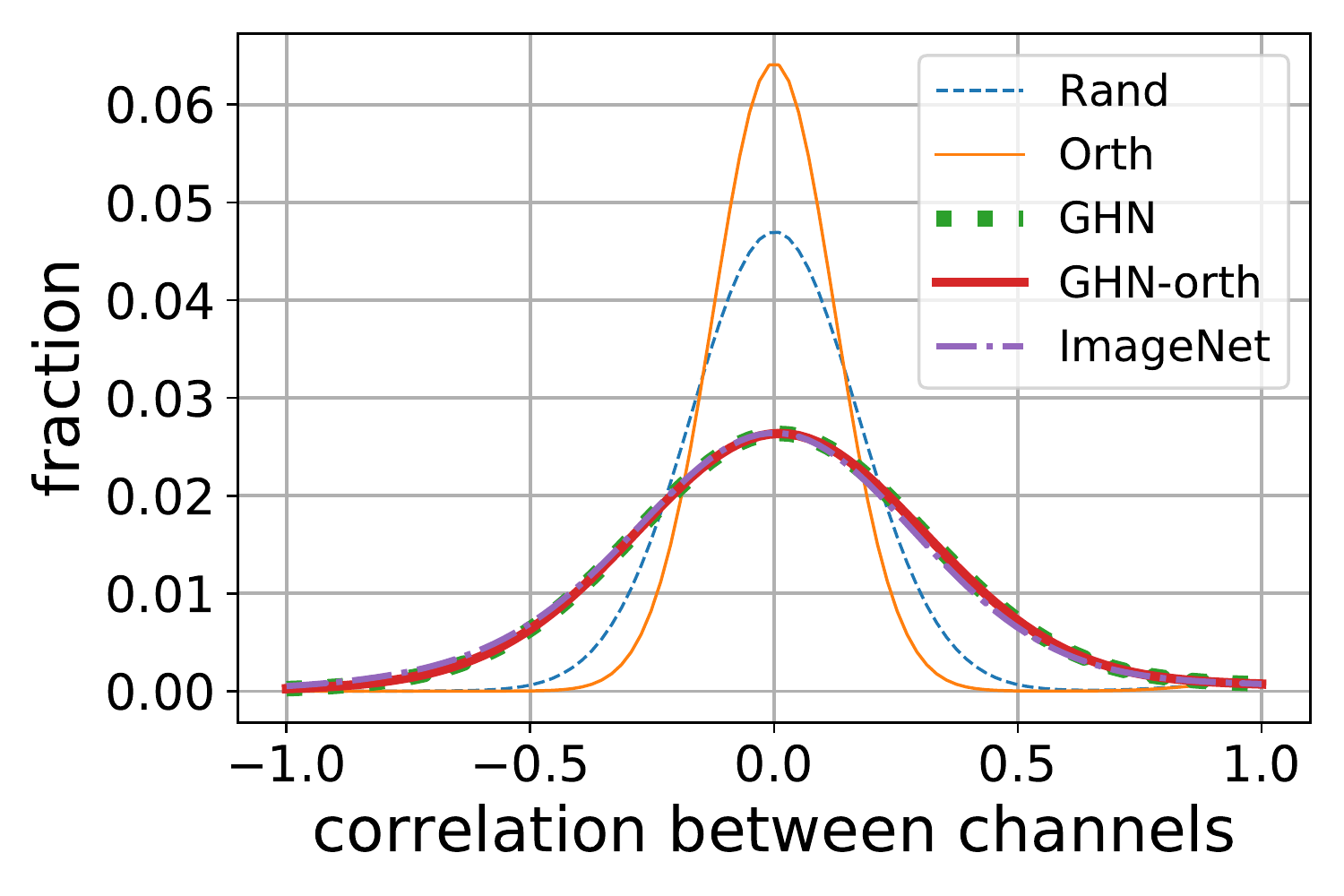}} & {\includegraphics[width=0.33\columnwidth]{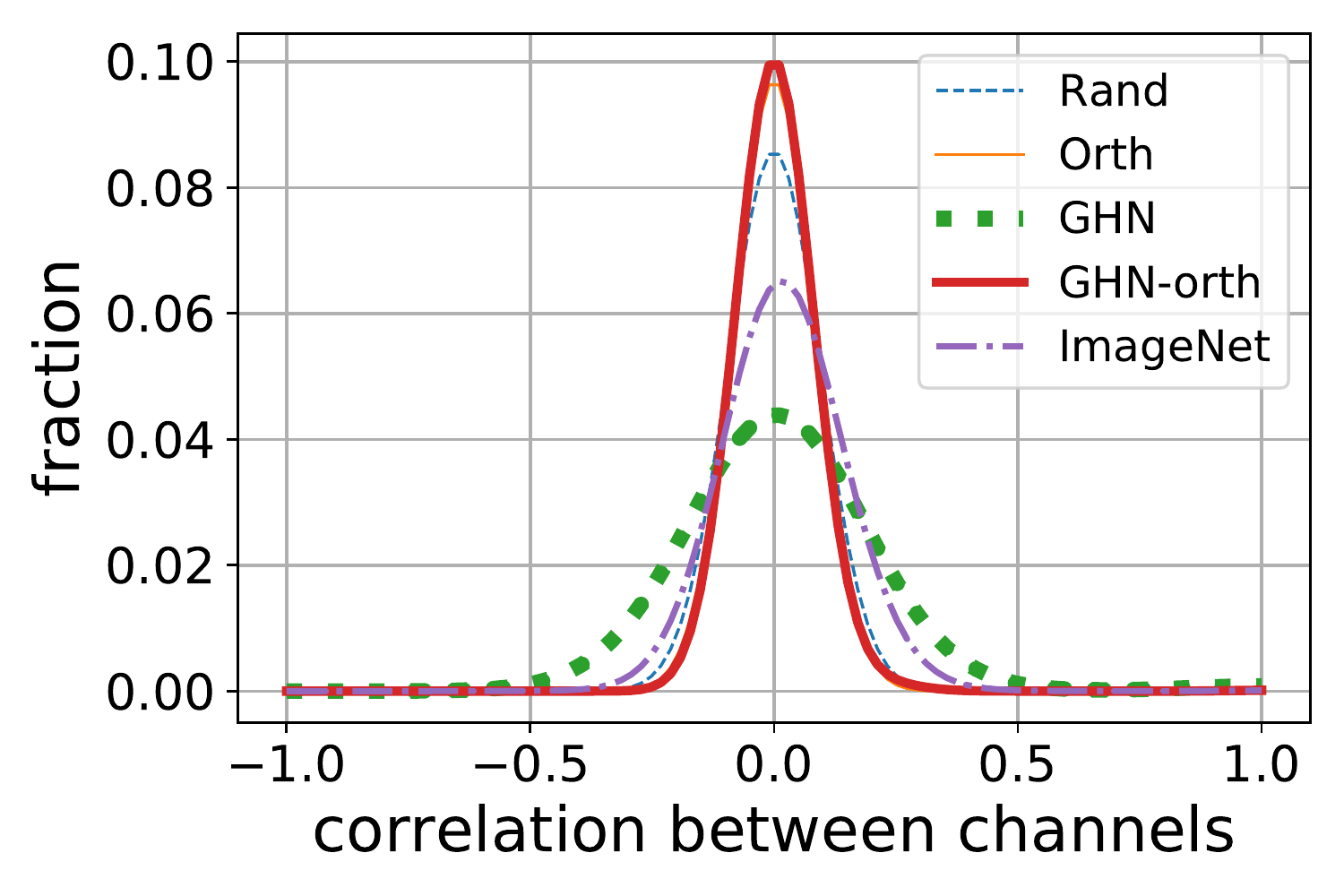}} &
         {\includegraphics[width=0.33\columnwidth]{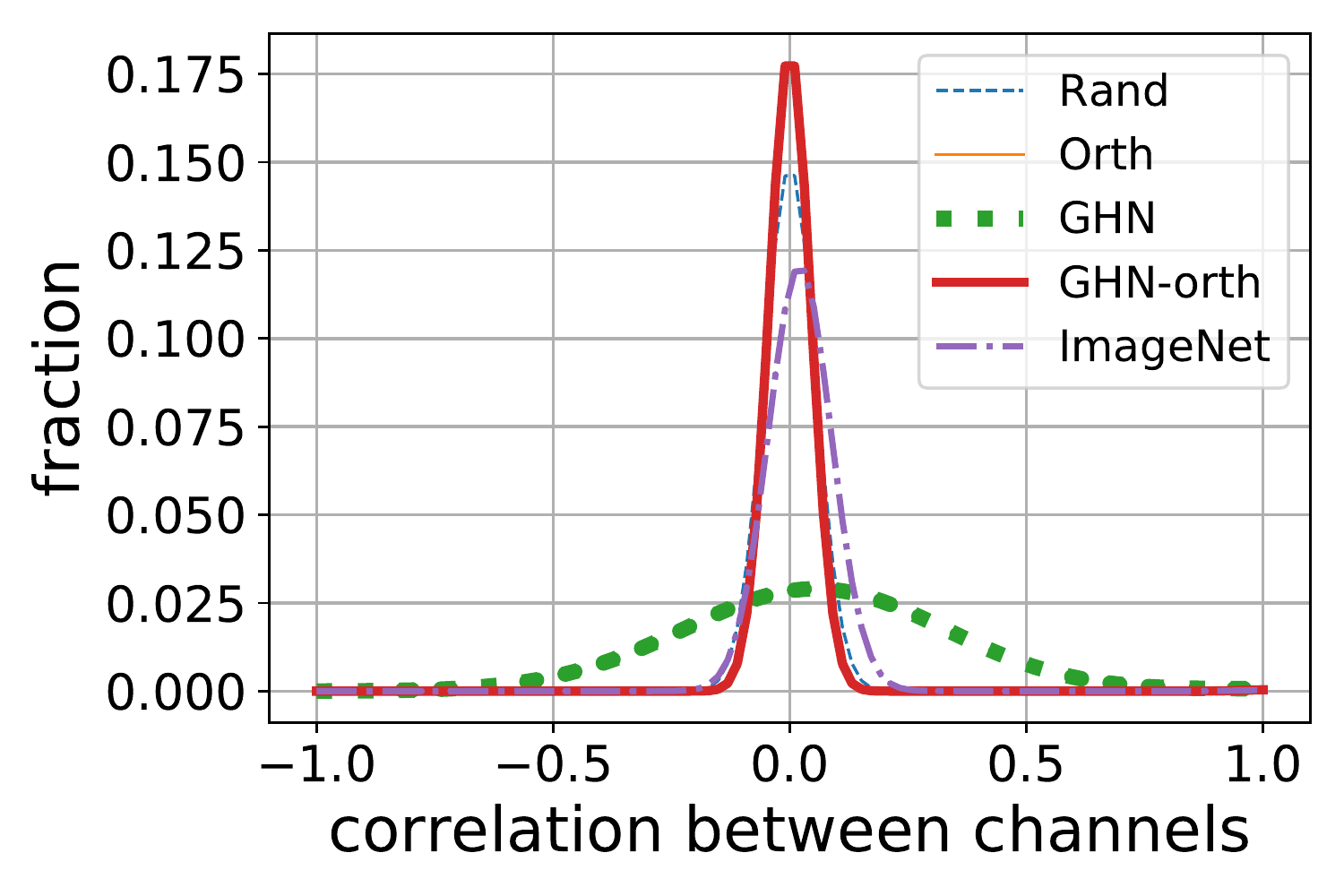}}  \\
         \multicolumn{3}{c}{{\includegraphics[width=0.85\columnwidth]{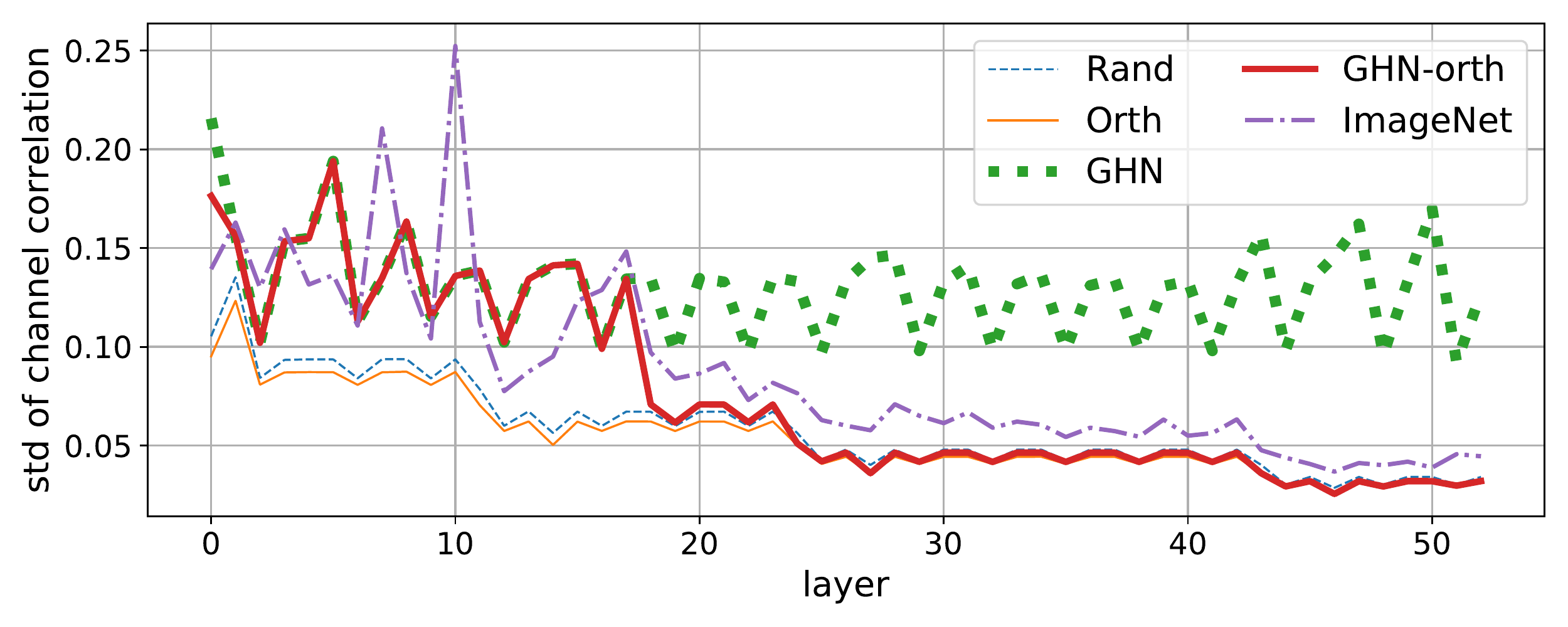}}}
\end{tabular}
\end{small}
\vspace{-10pt}
\caption{
The similarities between the channels of convolutional layers in ResNet-50 measured as Pearson's correlation. \textbf{Top}: distribution of correlation for selected layers. \textbf{Bottom}: standard deviation of correlation distribution for all layers. Lower correlation (around 0) is expected to be beneficial since it implies the parameters are linearly independent and, thus, can facilitate convergence and generalization~\cite{saxe2013exact,wang2020orthogonal}. We found that channels of the parameters predicted by Graph HyperNetworks (GHNs)~\cite{knyazev2021parameter} are highly correlated making their fine-tuning challenging. We alleviate this issue by post-processing (Section~\ref{sec:postproc}) predicted parameters (GHN-orth). We show the benefit of our post-processing in Section~\ref{sec:results}.
}
\label{fig:correlation}
\end{center}
\vskip -0.2in
\end{figure}

\section{Analysis of Parameters Predicted by GHNs}\label{sec:analyze}

Since the GHN predicts parameters from a highly-compressed low-dimensional representation~\cite{knyazev2021parameter}, we hypothesize that the predicted parameters may be highly correlated to each other.
The benefit of random initialization is that all parameters are drawn independently from some probability distribution, e.g. Gaussian: $\w_i \sim \cal{N}$, where $i$ is an index of the individual scalar value of the parameter tensor. In the GHN case, the parameters become conditional on a latent representation $\mathbf{z}$ of the input computational graph: $\w_i \sim p(\w_i | \mathbf{z})$.
To verify if the predicted parameters are highly correlated, we computed Pearson's correlation between channels for a given layer of a given architecture. We compared these correlations between networks with predicted parameters, initialized randomly and pretrained on ImageNet. We found that predicted parameters have generally much higher correlations with each other compared to other initializations (Figure~\ref{fig:correlation}). 
Methods such as random initialization and orthogonal regularization enforce statistical and linear independence of neural network parameters making them converge to a better solution in terms of generalization~\cite{arora2019fine,bansal2018can,wang2020orthogonal}.
As predicted parameters are highly correlated, their fine-tuning may be difficult with stochastic gradient descent and non-convex problems. We therefore propose to decorrelate predicted parameters without fully destroying their pretraining power.

\section{Post-processing of Predicted Parameters}\label{sec:postproc}

In a given neural net with the parameters predicted by GHNs, post-processing is performed for each $l$-th layer independently from other layers. Post-processing is performed only for convolutional and fully-connected  layers starting from a certain depth $L$, i.e. $l \geq L$, where $L$ is treated as a hyperparameter.
Post-processing of normalization layers and biases has been found to have no significant effect on the fine-tuning results.
We denote the parameters of the $l$-th layer as $\w_l$. 
Parameters of convolutional layers are 4D, $\w_l \in \mathbb{R}^{K \times C \times H \times W} $, while in certain post-processing steps a matrix (2D) form is required. Following~\cite{wang2020orthogonal}, to transform 4D to 2D, $\w_l$ is first reshaped to $\w_l \in \mathbb{R}^{K \times CHW} $ and then transposed if $K < CHW$. 
Post-processing consists of two steps: conditional noise addition (Section~\ref{sec:noise}) and orthogonal re-initialization (Section~\ref{sec:orth}).

\subsection{Conditional Noise Addition}\label{sec:noise}

In addition to the channels of parameters being highly correlated (Figure~\ref{fig:correlation}), we found that many parameters are identical, because the GHN of~\citet{knyazev2021parameter} copies the same tensor multiple times to make sure the shapes of the predicted and target parameters match.
Furthermore, the orthogonal re-initialization step introduced next in Section~\ref{sec:orth} can output less diverse parameters if the input parameters have a lot of identical values.
Therefore, to break the symmetry of identical parameters, we first add the Gaussian noise to the parameters in each $l$-th layer:
\begin{equation}
\label{eq:noise}
    \tilde{\w}_l = {\w}_l + \mathcal{N} \Big( 0, \beta\sigma(r({\w}_l)) \Big),
\end{equation}
\noindent where $r({\w}_l) \in \mathbb{R}^{K \times K} $ is the correlation between the channels of the parameters $\w_l$ (Figure~\ref{fig:correlation}), $\sigma(\cdot)$ is the standard deviation, while $\beta$ is a scaling factor shared across all layers. This way, the noise is added conditionally on the layer statistics to ensure that 
all layers are perturbed relatively equally.\looseness-1

\begin{figure*}[thbp]
\vskip 0.2in
\begin{center}
    \begin{tabular}{cc}
         {\includegraphics[width=0.42\textwidth,trim={3cm 3cm 3cm 3cm},clip]{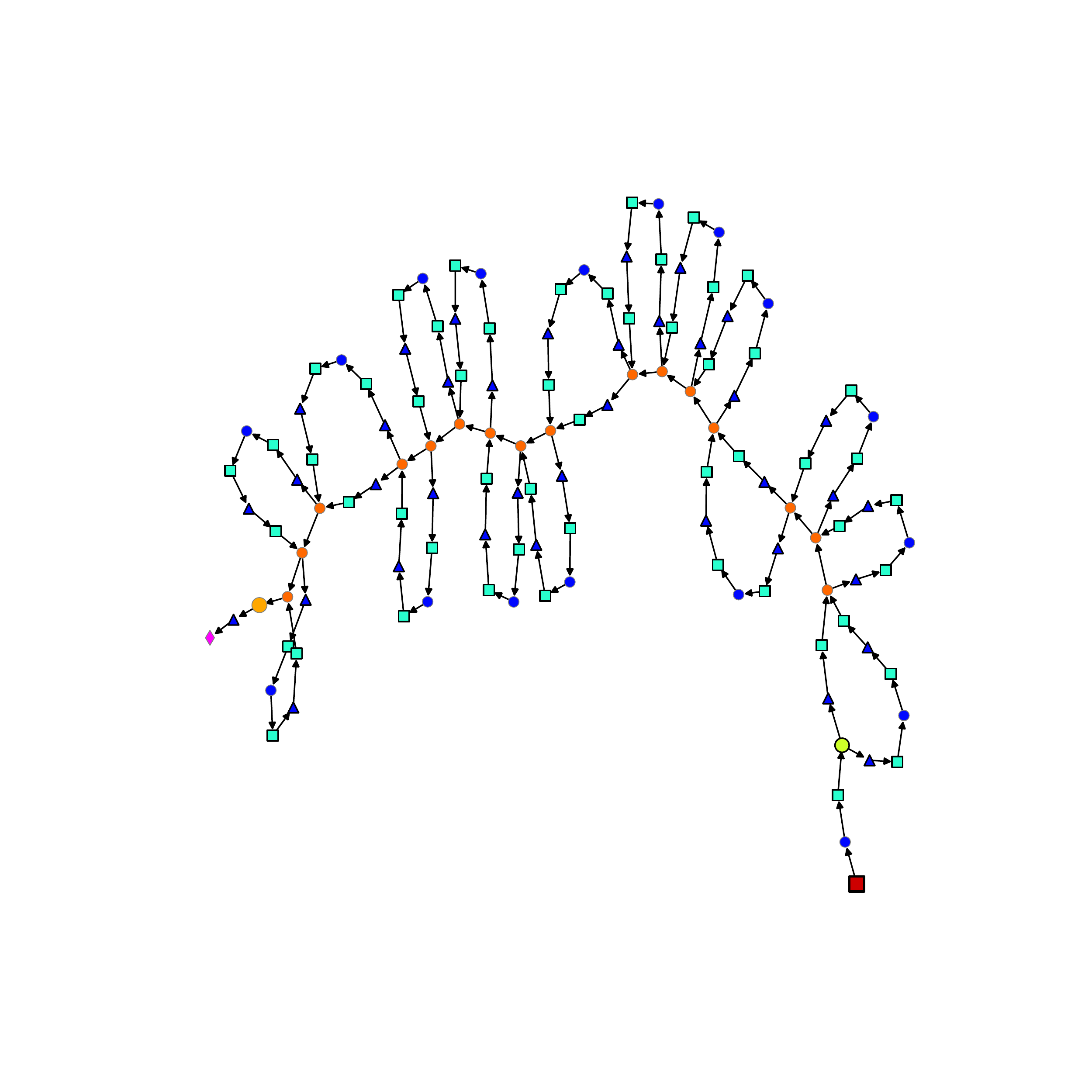}} & 
         {\includegraphics[width=0.42\textwidth,trim={3cm 3cm 3cm 3cm},clip]{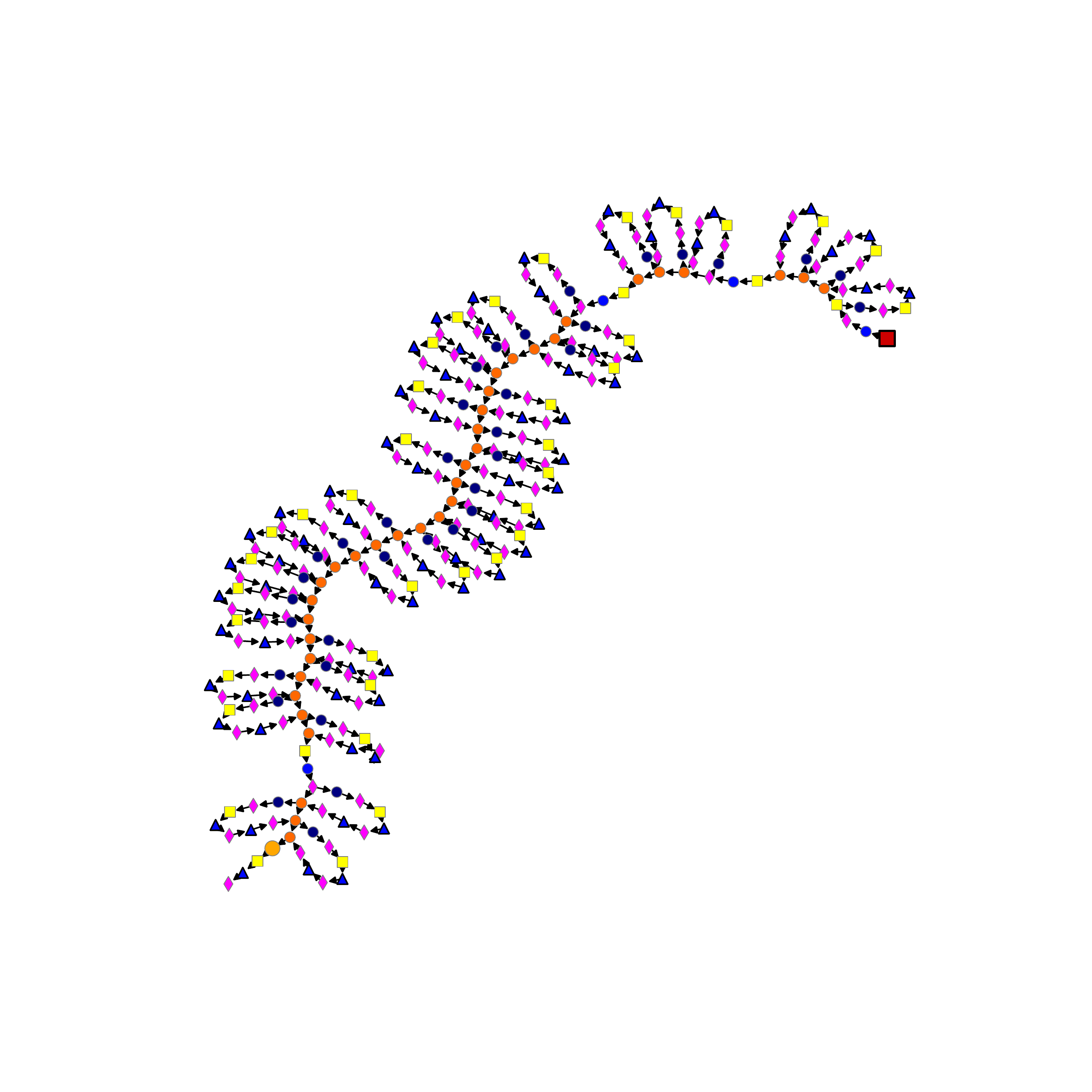}}
    \end{tabular}
    \vskip -0.2in
    \caption{Computational graphs of ResNet-50 (left) and ConvNeXt (right) used as inputs to the GHN to predict their parameters. Each node corresponds to a certain operation such as convolution, pooling, layer normalization, etc. For the detailed description of node shapes and colors see \cite{knyazev2021parameter}. }
    \label{fig:graphs}
\end{center}
\vskip -0.2in    
\end{figure*}

\subsection{Orthogonal Re-initialization}\label{sec:orth}

In the second step, we perform orthogonal re-initialization. 
We perform the same steps as in orthogonal initialization~\cite{saxe2013exact}\footnote{E.g. see the implementation in PyTorch~\cite{paszke2019pytorch}.}, but where initial parameters $\tilde{\w}_l$ are predicted by GHNs (with some noise added according to \eqref{eq:noise}) rather than drawn randomly from a Gaussian distribution.
Specifically, we first perform QR-decomposition $\tilde{\w}_l = \mathbf{Q} \mathbf{R}$ to find orthogonal matrix $\mathbf{Q}$ and upper-triangular matrix $\mathbf{R}$.
We then obtain new parameters $\tilde{\w}$ as following:
\begin{align}
\label{eq:qr}
    \tilde{\w}_l &= \mathbf{Q} \odot \text{sign} \Big( \text{diag}(\mathbf{R}) \Big).
\end{align}
Parameters $ \tilde{\w}_l$ are (transposed and) reshaped back to ${K \times C \times H \times W}$ and used instead of the original $\w_l$.

\section{Experimental Setup}\label{sec:setup}

We evaluate our parameter post-processing techniques on the fine-tuning task on CIFAR-10 image classification~\cite{krizhevsky2009learning} with 1000 (100 labels per class) labels as in~\cite{knyazev2021parameter}. 
For ImageNet-based and GHN-based initializations we replace the last classification layer and fine-tune all layers.
We fine-tune two neural nets: ResNet-50 and ConvNeXt (the base variant~\cite{liu2022convnet}) (Figure~\ref{fig:graphs}).
We identified five important hyperparameters: optimizer, number of epochs, initial learning rate, weight decay and input image size. 
The hyperparameters are selected from the following values:
\vspace{-10pt}
\begin{itemize}[leftmargin=5pt]
    \setlength\itemsep{0pt}
    \item optimizer: SGD, AdamW~\cite{loshchilov2017decoupled};
    \item number of epochs: 50, 100, 200, 300;
    \item initial learning rate: \{0.025, 0.01, 0.005, 0.0025, 0.001\} for SGD and \{0.004, 0.001, 0.0004, 0.0001\} for AdamW;
    \item weight decay: 0.0001, 0.001, 0.01, 0.05, 0.1;
    \item input image size: 224$\times$224, 32$\times$32 (original CIFAR-10 image size).
\end{itemize}

The batch size is fixed to 96 for ResNet-50 as in~\cite{zhang2018graph,knyazev2021parameter} and to 48 for ConvNeXt to fit into the memory of GPUs available to us. The cosine learning rate schedule is used in all experiments as in~\cite{zhang2018graph,knyazev2021parameter}.
For our method (\textsc{GHN-orth}), we have additional hyperparameters: layer $L$ from which to start post-processing (Section~\ref{sec:postproc}) and level of noise $\beta$ added to parameters in \eqref{eq:noise}.
 We tune all hyperparameters on the held-out validation set of 5,000 images.

\paragraph{Baselines}
As a baseline, we use random initialization~\cite{he2015delving} standard for ResNets, orthogonal initialization~\cite{saxe2013exact} and GHN-2 from~\cite{knyazev2021parameter} (denoted as GHN in this paper). Orthogonal initialization~\cite{saxe2013exact} is based on the same equation as \eqref{eq:qr}, but applied to the randomly-initialized parameters drawn from the Gaussian distribution.
As the oracle initialization we use ImageNet pretrained models.
For fair comparison, we tune hyperparameters the same way for all methods. The experiments are run five times with different random seeds. Mean and standard deviation of the accuracy on the test set of CIFAR-10 are reported in Table~\ref{tab:results}.

Experiments are done using the GHN code base of~\citet{knyazev2021parameter}: \url{https://github.com/facebookresearch/ppuda}.

\begin{table*}[thbp]
\caption{Classification accuracies (mean and standard deviation across 5 runs) on reduced CIFAR-10 with 1000 labels.}
\label{tab:results}
\vskip 0.05in
\begin{center}
\begin{small}
\begin{sc}
\begin{tabular}{lcc}
\toprule
\textbf{{Initialization}} & \textbf{ResNet-50} & \textbf{ConvNeXt} \\
& \cite{he2016deep} & \cite{liu2022convnet} \\
\# parameters & 23.5M & 87.6M \\
\midrule
Rand~\cite{he2015delving} & 60.6\std{0.5} & 48.0\std{0.9} \\
Orth~\cite{saxe2013exact} & 60.7\std{0.3} & 52.4\std{0.2} \\
GHN~\cite{knyazev2021parameter} & 61.8\std{0.3} & 52.3\std{0.5}\\
GHN-orth (ours) &  65.6\std{0.2} & 53.6\std{0.4} \\
ImageNet pretrained~\cite{he2016deep} & 88.7\std{0.2} & 95.6\std{0.1}  \\

\bottomrule
\end{tabular}
\end{sc}
\end{small}
\end{center}
\vskip -0.1in
\end{table*}

\section{Results}\label{sec:results}

\paragraph{Main results}
Our initialization based on post-processing predicted parameters (\textsc{GHN-orth}) improves on the direct competitor \textsc{GHN} by 3.8 and 1.3 absolute percentage points for ResNet-50 and ConvNeXt respectively (Table~\ref{tab:results}). These results demonstrate the importance of proposed parameter post-processing. \textsc{GHN-orth} also outperforms \textsc{Orth} confirming that our post-processing preserved useful structure in predicted parameters.
\textsc{GHN-orth} is significantly inferior to ImageNet-based initialization. However, \textsc{GHN-orth} takes only fractions of a second to initialize for ResNet-50, ConvNeXt and potentially many other upcoming neural architectures in the future.
In contrast, pretraining on ImageNet or large in-house datasets available to practitioners can take days or weeks for every novel architecture, especially given their increasing scale~\cite{zhai2022scaling}.

\paragraph{Other results}
Applying our post-processing steps to ImageNet-pretrained models have not been found helpful and reduced fine-tuning results (not reported in Table~\ref{tab:results}). This can be explained by the fact that the parameters of ImageNet-pretrained models are not highly correlated (Figure~\ref{fig:correlation}). While our post-processing can make them even more linearly and statistically independent, it can also damage high-quality filters.
As a sanity check, we also confirmed that applying \textsc{Orth} starting only from a specific layer $L$ the same way as in our \textsc{GHN-orth} (Section~\ref{sec:postproc}) was not better (60.2\std{0.8} for ResNet-50) than applying it to all layers (60.7\std{0.3}). This indicates that \textsc{GHN-orth} outperforms not simply due to tuning $L$. Ablated \textsc{GHN-orth} without adding the Gaussian noise \eqref{eq:noise} achieves 65.0\std{0.4} for ResNet-50, while without  orthogonal re-initialization \eqref{eq:qr} the performance drops to 61.4\std{0.2}. Therefore, both post-processing steps are important for the best performance.

\paragraph{Hyperparameters}
The best hyperparameters selected during tuning are reported in Table~\ref{tab:hyperparams}.
\textsc{GHN} and \textsc{GHN-orth} have consistently the same hyperparameters that are \textit{slightly} different from the hyperparameters of \textsc{Rand} and \textsc{Orth} and \textit{very} different from the best hyperparameters of ImageNet pretrained models.
For example, despite the GHN was trained on 224$\times$224 images of ImageNet, fine-tuning on 32$\times$32 yielded better results for both \textsc{GHN} and \textsc{GHN-orth}.
In that sense, predicted parameters are closer to random-based initialization methods.

\begin{figure*}[thbp]
\begin{center}
\begin{tabular}{cc}
    \textsc{ResNet-50} & \textsc{ConvNeXt} \\
     \includegraphics[width=0.7\columnwidth]{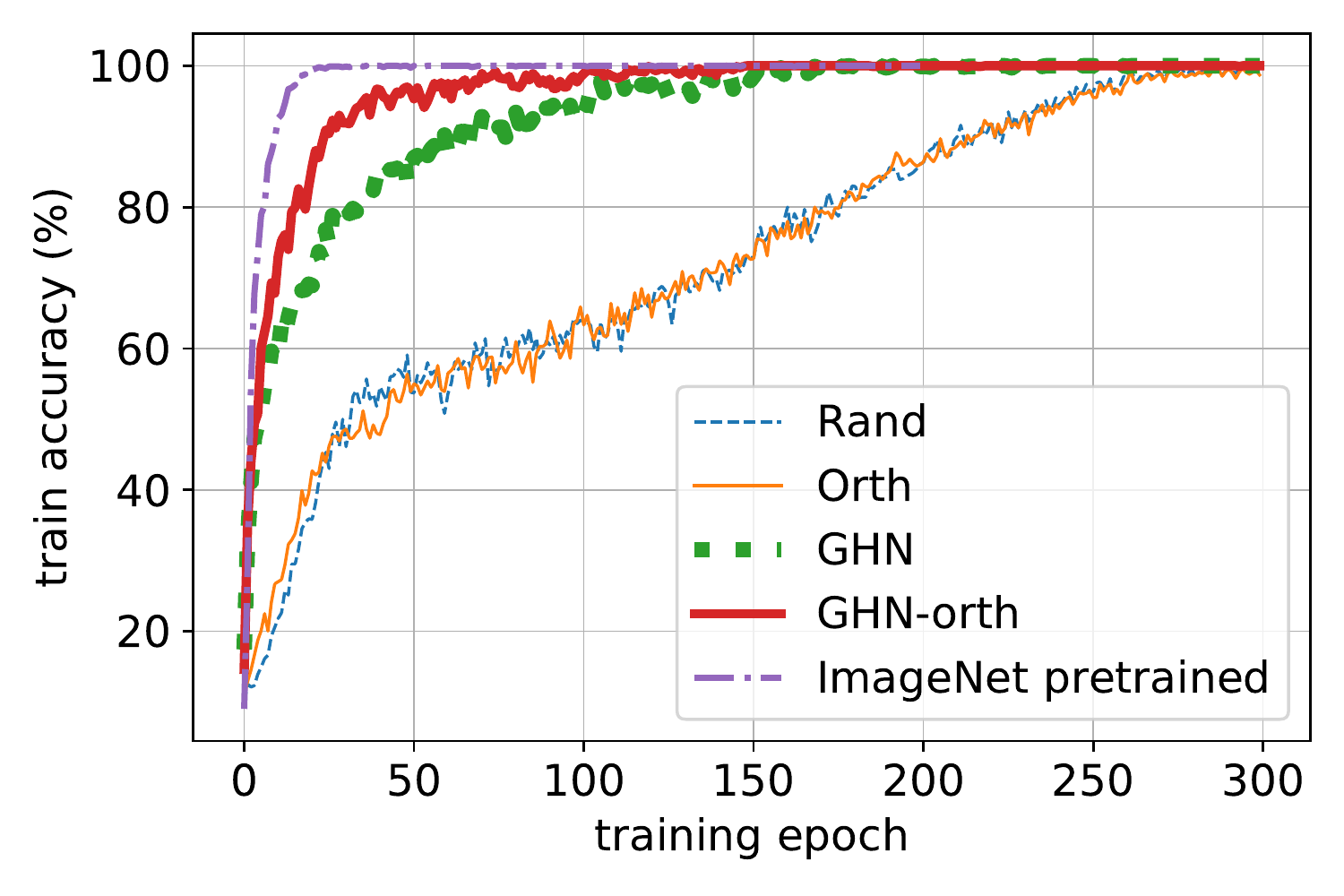} & \includegraphics[width=0.7\columnwidth]{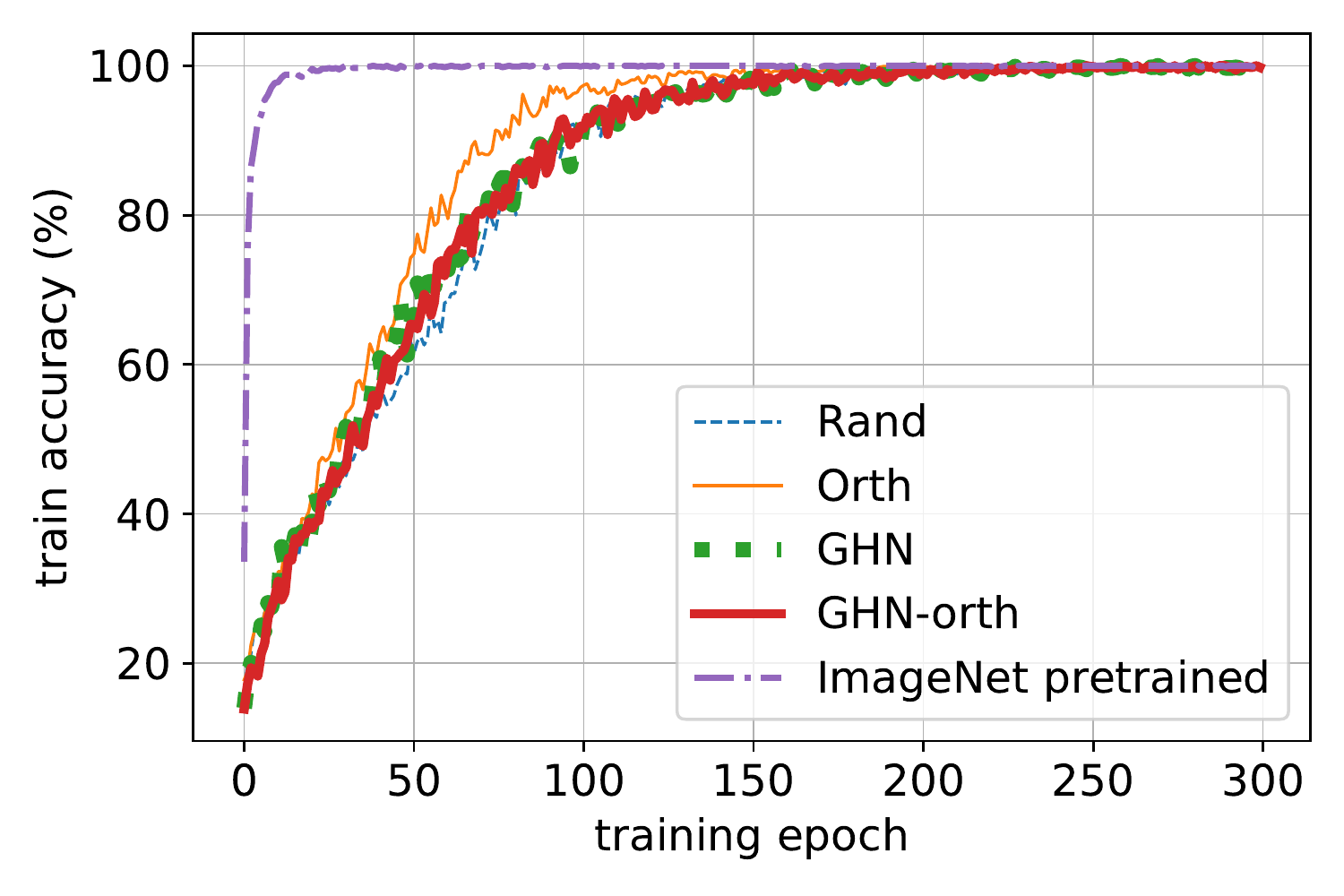} \\ \includegraphics[width=0.7\columnwidth]{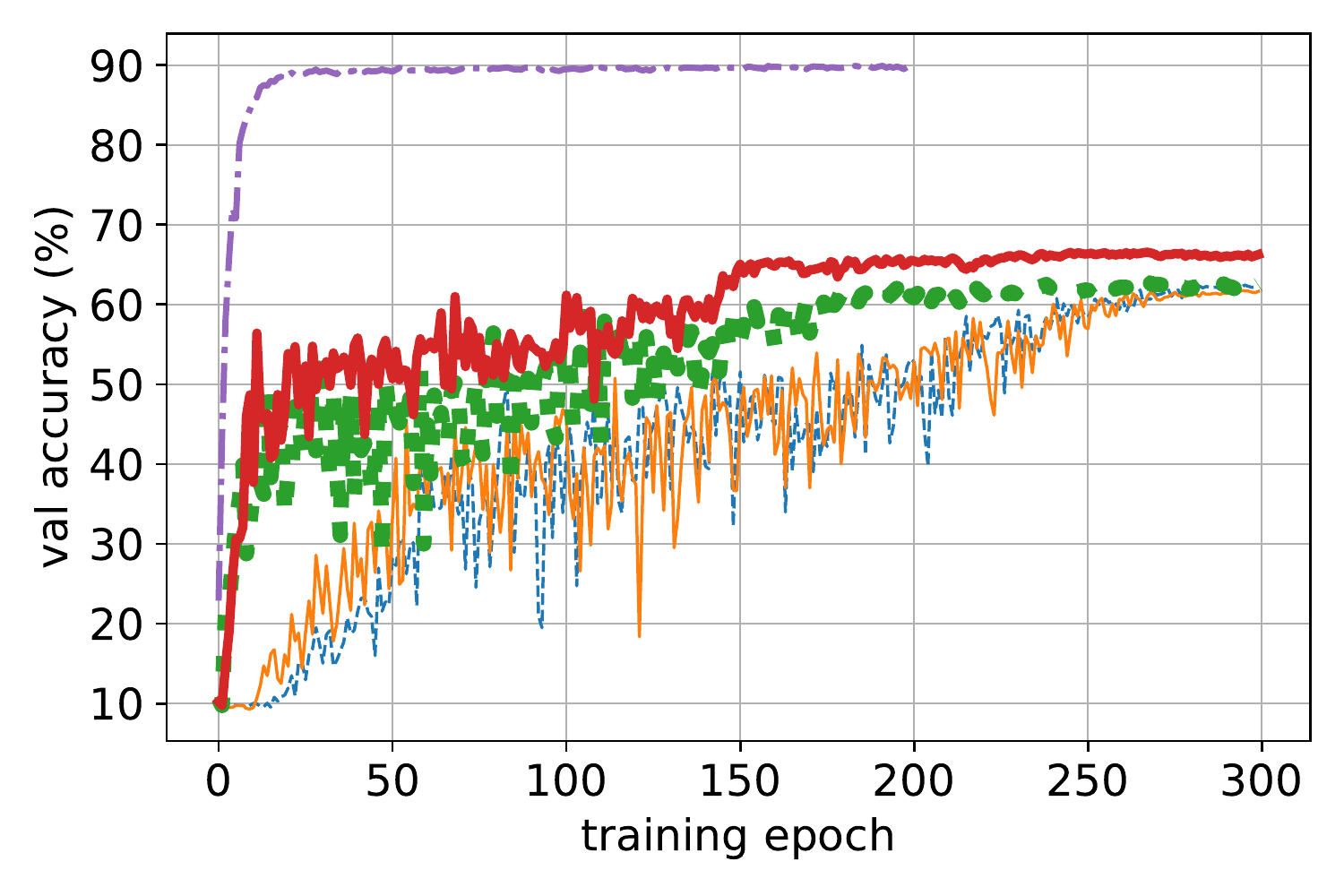}
      & \includegraphics[width=0.7\columnwidth]{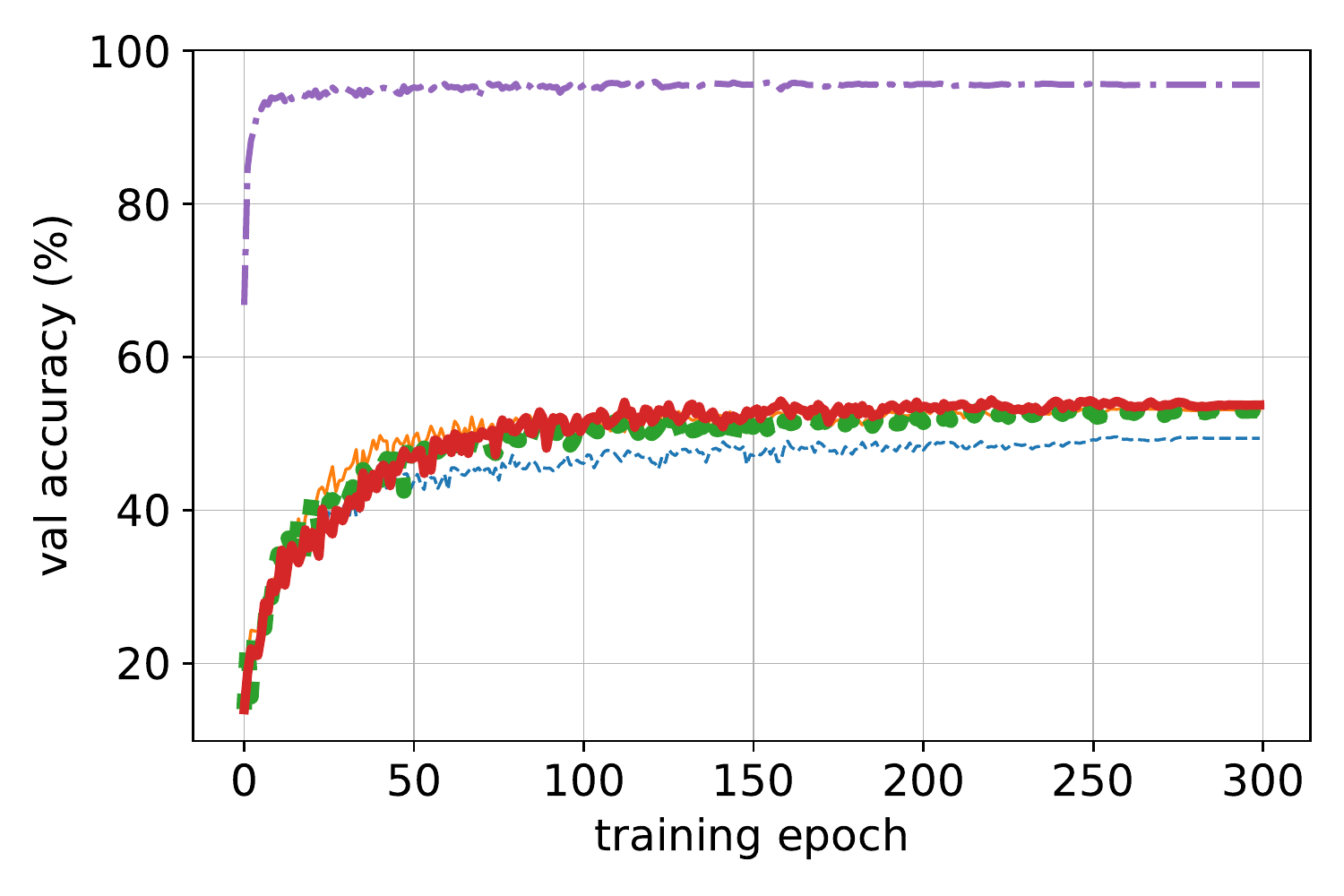}
\end{tabular}
\vskip -0.2in
\caption{Training (top) and validation (bottom) accuracies on CIFAR-10 with 1000 labels for ResNet-50 (left) and ConvNeXt (right).}
\label{fig:curves}
\end{center}
\vskip -0.2in
\end{figure*}

\paragraph{Training curves}
Fine-tuning both ResNet-50 and ConvNeXt for longer ($\geq$ 200 epochs) turned out to be beneficial in all cases as long as other hyperparameters are adjusted accordingly (Table~\ref{tab:hyperparams}). However, ImageNet pretrained models converge to strong performance extremely fast (in 10-20 epochs) compared to other methods (Figure~\ref{fig:curves}). 
Our \textsc{GHN-orth} also allows ResNet-50 to converge faster compared to other methods. At the same time, for ConvNeXt there is no significant convergence benefit of using our method even though the final classification performance is better. 
For ConvNeXt, training and validation accuracies do not increase very fast (except for \textsc{ImageNet pretrained}) indicating that training for more than 300 epochs may be useful.\looseness-1

\begin{table}[thbp]
\vskip-0.05in
\caption{Hyperparameters selected during tuning and used to train models in Table~\ref{tab:results}. The hyperparameters are in the following order: optimizer; number of epochs; initial learning rate; weight decay; input image size; noise scaling factor $\beta$; layer $L$ (see Section~\ref{sec:postproc}). }
\label{tab:hyperparams}
\vskip-0.05in
\begin{center}
\begin{tiny}
\begin{sc}
\setlength{\tabcolsep}{1pt}
\begin{tabular}{l|l|l}
\toprule
\textbf{{Init}} & \multicolumn{1}{c|}{\textbf{ResNet-50}} & \multicolumn{1}{c}{\textbf{ConvNeXt}} \\
batch size & \multicolumn{1}{c|}{96} & \multicolumn{1}{c}{48} \\
\midrule

Rand & SGD; 300; 0.01; 0.05; 32 & AdamW; 300; 0.001; 1e-4; 32 \\

Orth & SGD; 300; 0.01; 0.05; 32 & AdamW; 300; 0.001; 1e-4; 32 \\

GHN & SGD; 300; 0.01; 0.01; 32 & AdamW; 300; 0.001; 0.1; 32\\

GHN-orth &  SGD; 300; 0.01; 0.01; 32; 3e-5; 37 & AdamW; 300; 0.001; 0.1; 32; 3e-5; 100 \\

ImageNet & SGD; 200; 0.001; 1e-4; 224 & SGD; 300; 0.0025; 1e-4; 224  \\

\bottomrule
\end{tabular}
\end{sc}
\end{tiny}
\end{center}
\vskip -0.1in
\end{table}

\section{Discussion}\label{sec:discuss}

\begin{figure}[thbp]
\vspace{-10pt}
\begin{center}
    \includegraphics[width=\columnwidth]{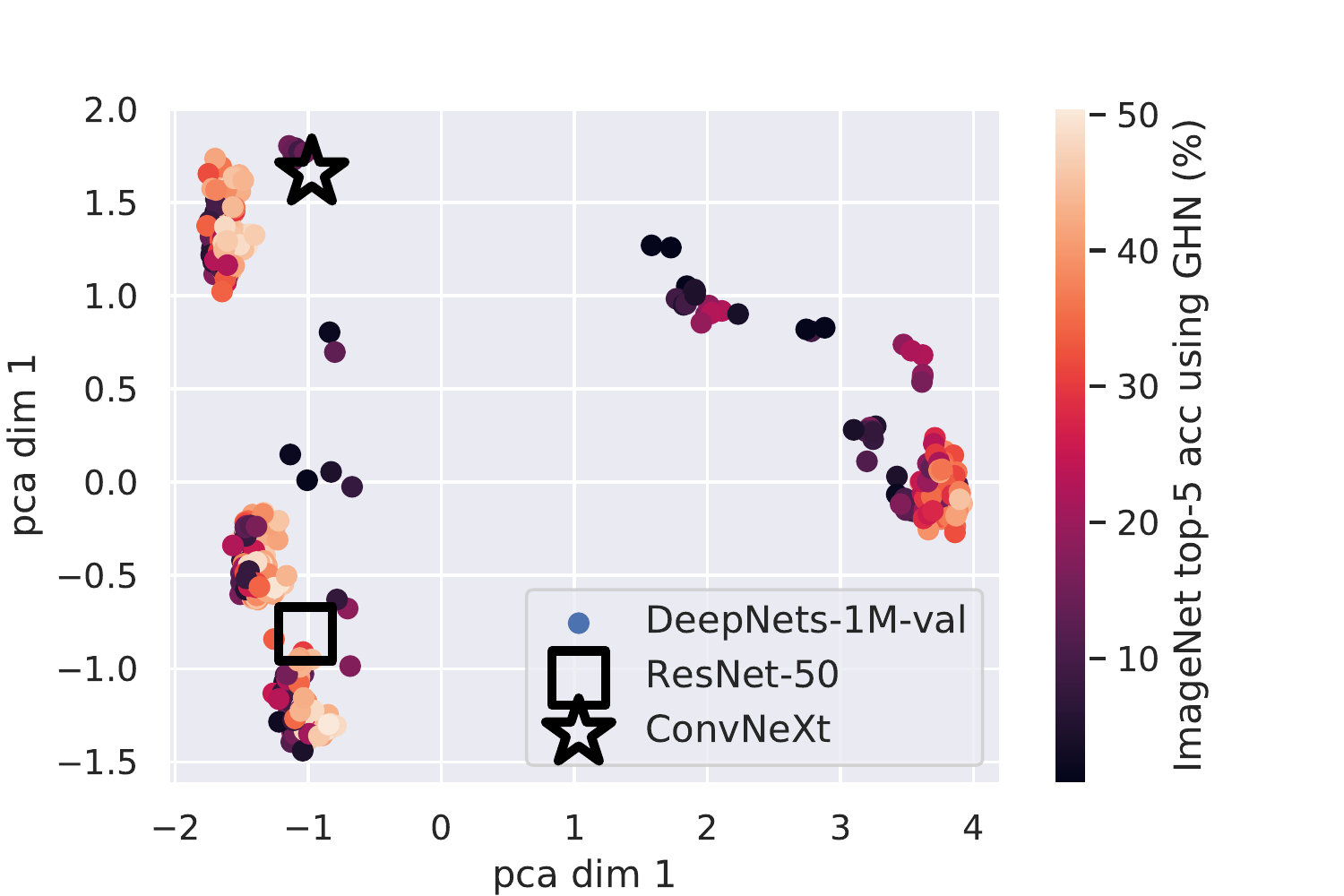}
    \vspace{-18pt}
    \caption{PCA-based projection of 32-dimensional latent representations of input computational graphs to two dimensionalities. These representations are computed based on~\cite{knyazev2021parameter} on 500 validation architectures of DeepNets-1M. }
    \label{fig:pca}
\end{center}
\vskip -0.2in
\end{figure}

The benefit of \textsc{GHN-orth} and \textsc{GHN} is lower on ConvNeXt than on ResNet-50 (Table~\ref{tab:results}). The capacities of these architectures (in terms of the number of trainable parameters) are not that different to explain this difference. We argue that even though ConvNeXt is composed of largely the same primitive operations\footnote{One of the operations used in ConvNeXt was not supported by the GHNs, so we did not predict their parameters, which accounted for a small percentage w.r.t. the total number of parameters in ConvNeXt. There are also some operations without trainable parameters such as GELU nonlinearities or permutation of dimensions that are not explicitly modeled by GHNs and not included in the input computational graphs.} that compose the training architectures of GHNs (DeepNets-1M~\cite{knyazev2021parameter}), the compositions of these primitives in ConvNeXt are quite different compared to the other architectures in DeepNets-1M and ResNet-50. Such a difference in compositions may create a significant distribution shift confusing the GHN and making it predict poor parameters. To visualize this effect, we first extracted latent representations of input computational graphs of the validation architectures of DeepNets-1M as well as of ResNet-50 and ConvNeXt the same way as in~\cite{knyazev2021parameter}. We then projected these representations using the principal component analysis (PCA) into two dimensionalities and color coded the architectures with the accuracies of the corresponding networks with predicted parameters (Figure~\ref{fig:pca}). This visualization reveals distinct clusters for lower and higher performant architectures in GHN's latent space. While ResNet-50 is located closely to the clusters with higher performant architectures, ConvNeXt is grouped together with a few low performant architectures.
A relatively outlying latent representation of ConvNeXt may be explained by either the lack of similar architectures in the training set of DeepNets-1M or due to the difficulty of training the GHN on this kind of architecture. Understanding these reasons better may lead to more advances in GHNs and may potentially bridge the gap between computationally-intensive pretraining of networks with SGD and almost zero-cost parameter prediction in the transfer learning scenarios.

Alternative to our approach, efficient pretraining of large networks is possible by first pretraining a smaller version of the network and then growing it~\cite{chen2015net2net,evci2022gradmax}. However, parameter prediction using GHNs is even more efficient (assuming a trained GHN already exists) as it does not require pretraining networks.
We also have not compared our approach to many other advanced initialization methods such as~\cite{mishkin2015all,knyazev2017recursive,zhang2019fixup,huang2020improving,zhang2019fixup,dauphin2019metainit,zhu2021gradinit,elsken2020meta}, which is left for future work. 

\section*{Acknowledgements}
We would like to thank Diganta Misra, Bharat Runwal, Marwa El Halabi, Yan Zhang and Graham Taylor for the useful discussion and feedback. 
Resources used in preparing this research were provided by Calcul Québec (www.calculquebec.ca), Compute Canada (www.computecanada.ca) and Machine Learning Research Group at the University of Guelph (www.gwtaylor.ca).

\bibliographystyle{icml2022}
\bibliography{ref}

\begin{thebibliography}{28}
\providecommand{\natexlab}[1]{#1}
\providecommand{\url}[1]{\texttt{#1}}
\expandafter\ifx\csname urlstyle\endcsname\relax
  \providecommand{\doi}[1]{doi: #1}\else
  \providecommand{\doi}{doi: \begingroup \urlstyle{rm}\Url}\fi

\bibitem[Arora et~al.(2019)Arora, Du, Hu, Li, and Wang]{arora2019fine}
Arora, S., Du, S., Hu, W., Li, Z., and Wang, R.
\newblock Fine-grained analysis of optimization and generalization for
  overparameterized two-layer neural networks.
\newblock In \emph{International Conference on Machine Learning}, pp.\
  322--332. PMLR, 2019.

\bibitem[Bansal et~al.(2018)Bansal, Chen, and Wang]{bansal2018can}
Bansal, N., Chen, X., and Wang, Z.
\newblock Can we gain more from orthogonality regularizations in training deep
  networks?
\newblock \emph{Advances in Neural Information Processing Systems}, 31, 2018.

\bibitem[Chen et~al.(2015)Chen, Goodfellow, and Shlens]{chen2015net2net}
Chen, T., Goodfellow, I., and Shlens, J.
\newblock Net2net: Accelerating learning via knowledge transfer.
\newblock \emph{arXiv preprint arXiv:1511.05641}, 2015.

\bibitem[Dauphin \& Schoenholz(2019)Dauphin and
  Schoenholz]{dauphin2019metainit}
Dauphin, Y. and Schoenholz, S.~S.
\newblock Metainit: Initializing learning by learning to initialize.
\newblock 2019.

\bibitem[Dosovitskiy et~al.(2020)Dosovitskiy, Beyer, Kolesnikov, Weissenborn,
  Zhai, Unterthiner, Dehghani, Minderer, Heigold, Gelly,
  et~al.]{dosovitskiy2020image}
Dosovitskiy, A., Beyer, L., Kolesnikov, A., Weissenborn, D., Zhai, X.,
  Unterthiner, T., Dehghani, M., Minderer, M., Heigold, G., Gelly, S., et~al.
\newblock An image is worth 16x16 words: Transformers for image recognition at
  scale.
\newblock \emph{arXiv preprint arXiv:2010.11929}, 2020.

\bibitem[Elsken et~al.(2019)Elsken, Metzen, and Hutter]{elsken2019neural}
Elsken, T., Metzen, J.~H., and Hutter, F.
\newblock Neural architecture search: A survey.
\newblock \emph{The Journal of Machine Learning Research}, 20\penalty0
  (1):\penalty0 1997--2017, 2019.

\bibitem[Elsken et~al.(2020)Elsken, Staffler, Metzen, and
  Hutter]{elsken2020meta}
Elsken, T., Staffler, B., Metzen, J.~H., and Hutter, F.
\newblock Meta-learning of neural architectures for few-shot learning.
\newblock In \emph{Proceedings of the IEEE/CVF conference on computer vision
  and pattern recognition}, pp.\  12365--12375, 2020.

\bibitem[Evci et~al.(2022)Evci, Vladymyrov, Unterthiner, van Merri{\"e}nboer,
  and Pedregosa]{evci2022gradmax}
Evci, U., Vladymyrov, M., Unterthiner, T., van Merri{\"e}nboer, B., and
  Pedregosa, F.
\newblock Gradmax: Growing neural networks using gradient information.
\newblock \emph{arXiv preprint arXiv:2201.05125}, 2022.

\bibitem[Glorot \& Bengio(2010)Glorot and Bengio]{glorot2010understanding}
Glorot, X. and Bengio, Y.
\newblock Understanding the difficulty of training deep feedforward neural
  networks.
\newblock In \emph{Proceedings of the thirteenth international conference on
  artificial intelligence and statistics}, pp.\  249--256, 2010.

\bibitem[He et~al.(2015)He, Zhang, Ren, and Sun]{he2015delving}
He, K., Zhang, X., Ren, S., and Sun, J.
\newblock Delving deep into rectifiers: Surpassing human-level performance on
  imagenet classification.
\newblock In \emph{Proceedings of the IEEE international conference on computer
  vision}, pp.\  1026--1034, 2015.

\bibitem[He et~al.(2016)He, Zhang, Ren, and Sun]{he2016deep}
He, K., Zhang, X., Ren, S., and Sun, J.
\newblock Deep residual learning for image recognition.
\newblock In \emph{Proceedings of the IEEE conference on computer vision and
  pattern recognition}, pp.\  770--778, 2016.

\bibitem[Huang et~al.(2020)Huang, Perez, Ba, and Volkovs]{huang2020improving}
Huang, X.~S., Perez, F., Ba, J., and Volkovs, M.
\newblock Improving transformer optimization through better initialization.
\newblock In \emph{International Conference on Machine Learning}, pp.\
  4475--4483. PMLR, 2020.

\bibitem[Huh et~al.(2016)Huh, Agrawal, and Efros]{huh2016makes}
Huh, M., Agrawal, P., and Efros, A.~A.
\newblock What makes imagenet good for transfer learning?
\newblock \emph{arXiv preprint arXiv:1608.08614}, 2016.

\bibitem[Knyazev et~al.(2017)Knyazev, Barth, and
  Martinetz]{knyazev2017recursive}
Knyazev, B., Barth, E., and Martinetz, T.
\newblock Recursive autoconvolution for unsupervised learning of convolutional
  neural networks.
\newblock In \emph{2017 International Joint Conference on Neural Networks
  (IJCNN)}, pp.\  2486--2493. IEEE, 2017.

\bibitem[Knyazev et~al.(2021)Knyazev, Drozdzal, Taylor, and
  Romero~Soriano]{knyazev2021parameter}
Knyazev, B., Drozdzal, M., Taylor, G.~W., and Romero~Soriano, A.
\newblock Parameter prediction for unseen deep architectures.
\newblock \emph{Advances in Neural Information Processing Systems}, 34, 2021.

\bibitem[Kolesnikov et~al.(2020)Kolesnikov, Beyer, Zhai, Puigcerver, Yung,
  Gelly, and Houlsby]{kolesnikov2020big}
Kolesnikov, A., Beyer, L., Zhai, X., Puigcerver, J., Yung, J., Gelly, S., and
  Houlsby, N.
\newblock Big transfer (bit): General visual representation learning.
\newblock In \emph{European conference on computer vision}, pp.\  491--507.
  Springer, 2020.

\bibitem[Krizhevsky et~al.(2009)]{krizhevsky2009learning}
Krizhevsky, A. et~al.
\newblock Learning multiple layers of features from tiny images.
\newblock 2009.

\bibitem[Liu et~al.(2022)Liu, Mao, Wu, Feichtenhofer, Darrell, and
  Xie]{liu2022convnet}
Liu, Z., Mao, H., Wu, C.-Y., Feichtenhofer, C., Darrell, T., and Xie, S.
\newblock A convnet for the 2020s.
\newblock \emph{arXiv preprint arXiv:2201.03545}, 2022.

\bibitem[Loshchilov \& Hutter(2017)Loshchilov and
  Hutter]{loshchilov2017decoupled}
Loshchilov, I. and Hutter, F.
\newblock Decoupled weight decay regularization.
\newblock \emph{arXiv preprint arXiv:1711.05101}, 2017.

\bibitem[Mishkin \& Matas(2015)Mishkin and Matas]{mishkin2015all}
Mishkin, D. and Matas, J.
\newblock All you need is a good init.
\newblock \emph{arXiv preprint arXiv:1511.06422}, 2015.

\bibitem[Paszke et~al.(2019)Paszke, Gross, Massa, Lerer, Bradbury, Chanan,
  Killeen, Lin, Gimelshein, Antiga, et~al.]{paszke2019pytorch}
Paszke, A., Gross, S., Massa, F., Lerer, A., Bradbury, J., Chanan, G., Killeen,
  T., Lin, Z., Gimelshein, N., Antiga, L., et~al.
\newblock Pytorch: An imperative style, high-performance deep learning library.
\newblock \emph{Advances in neural information processing systems}, 32, 2019.

\bibitem[Russakovsky et~al.(2015)Russakovsky, Deng, Su, Krause, Satheesh, Ma,
  Huang, Karpathy, Khosla, Bernstein, et~al.]{russakovsky2015imagenet}
Russakovsky, O., Deng, J., Su, H., Krause, J., Satheesh, S., Ma, S., Huang, Z.,
  Karpathy, A., Khosla, A., Bernstein, M., et~al.
\newblock Imagenet large scale visual recognition challenge.
\newblock \emph{International journal of computer vision}, 115\penalty0
  (3):\penalty0 211--252, 2015.

\bibitem[Saxe et~al.(2013)Saxe, McClelland, and Ganguli]{saxe2013exact}
Saxe, A.~M., McClelland, J.~L., and Ganguli, S.
\newblock Exact solutions to the nonlinear dynamics of learning in deep linear
  neural networks.
\newblock \emph{arXiv preprint arXiv:1312.6120}, 2013.

\bibitem[Wang et~al.(2020)Wang, Chen, Chakraborty, and Yu]{wang2020orthogonal}
Wang, J., Chen, Y., Chakraborty, R., and Yu, S.~X.
\newblock Orthogonal convolutional neural networks.
\newblock In \emph{Proceedings of the IEEE/CVF conference on computer vision
  and pattern recognition}, pp.\  11505--11515, 2020.

\bibitem[Zhai et~al.(2022)Zhai, Kolesnikov, Houlsby, and
  Beyer]{zhai2022scaling}
Zhai, X., Kolesnikov, A., Houlsby, N., and Beyer, L.
\newblock Scaling vision transformers.
\newblock In \emph{Proceedings of the IEEE/CVF Conference on Computer Vision
  and Pattern Recognition}, pp.\  12104--12113, 2022.

\bibitem[Zhang et~al.(2018)Zhang, Ren, and Urtasun]{zhang2018graph}
Zhang, C., Ren, M., and Urtasun, R.
\newblock Graph hypernetworks for neural architecture search.
\newblock \emph{arXiv preprint arXiv:1810.05749}, 2018.

\bibitem[Zhang et~al.(2019)Zhang, Dauphin, and Ma]{zhang2019fixup}
Zhang, H., Dauphin, Y.~N., and Ma, T.
\newblock Fixup initialization: Residual learning without normalization.
\newblock \emph{arXiv preprint arXiv:1901.09321}, 2019.

\bibitem[Zhu et~al.(2021)Zhu, Ni, Xu, Kong, Huang, and
  Goldstein]{zhu2021gradinit}
Zhu, C., Ni, R., Xu, Z., Kong, K., Huang, W.~R., and Goldstein, T.
\newblock Gradinit: Learning to initialize neural networks for stable and
  efficient training.
\newblock \emph{arXiv preprint arXiv:2102.08098}, 2021.

\end{thebibliography}


\end{document}